\documentclass[sigconf]{acmart}

\usepackage{textcomp}
\usepackage{multirow}
\usepackage[T1]{fontenc}
\usepackage[english]{babel}
\usepackage{microtype}
\AtBeginDocument{%
  }

\copyrightyear{2026}
\acmYear{2026}
\setcopyright{cc}
\setcctype{by}

\acmConference[UIST Adjunct '26]%
  {The 39th Annual ACM Symposium on User Interface Software and Technology}%
  {November 02--05, 2026}%
  {Detroit, MI, USA}

\acmBooktitle{The 39th Annual ACM Symposium on User Interface Software and Technology (UIST Adjunct '26), November 02--05, 2026, Detroit, MI, USA}

\acmDOI{10.1145/3830397.3841893}
\acmISBN{979-8-4007-2855-6/2026/11}

\begin{document}

\title{Warping the Workspace: Expanding Visual Access with Adjustable Reach Mapping for Humanoid Teleoperation}

\author{Keigo Minamida}
\authornote{Both authors contributed equally to this work.}
\affiliation{%
  \institution{The University of Tokyo} 
  \institution{Sony CSL - Kyoto}
  \city{Tokyo, Kyoto}
  \country{Japan}
}
\email{keigo-minamida@g.ecc.u-tokyo.ac.jp}

\author{Koya Dendo}
\authornotemark[1]
\affiliation{%
  \institution{The University of Tokyo}
  \institution{Sony CSL - Kyoto}
  \city{Tokyo, Kyoto}
  \country{Japan}
}
\email{koya-6732@g.ecc.u-tokyo.ac.jp}

\author{Yuto Nakamura}
\affiliation{%
  \institution{The University of Tokyo}
  \city{Tokyo}
  \country{Japan}}
\email{yuto-nakamura@g.ecc.u-tokyo.ac.jp}

\author{Jun Rekimoto}
\affiliation{
  \institution{Sony CSL - Kyoto}
  \city{Kyoto}
  \country{Japan}
}
\email{rekimoto@acm.org}



\begin{teaserfigure}
  \centering
  \includegraphics[width=.94\textwidth]{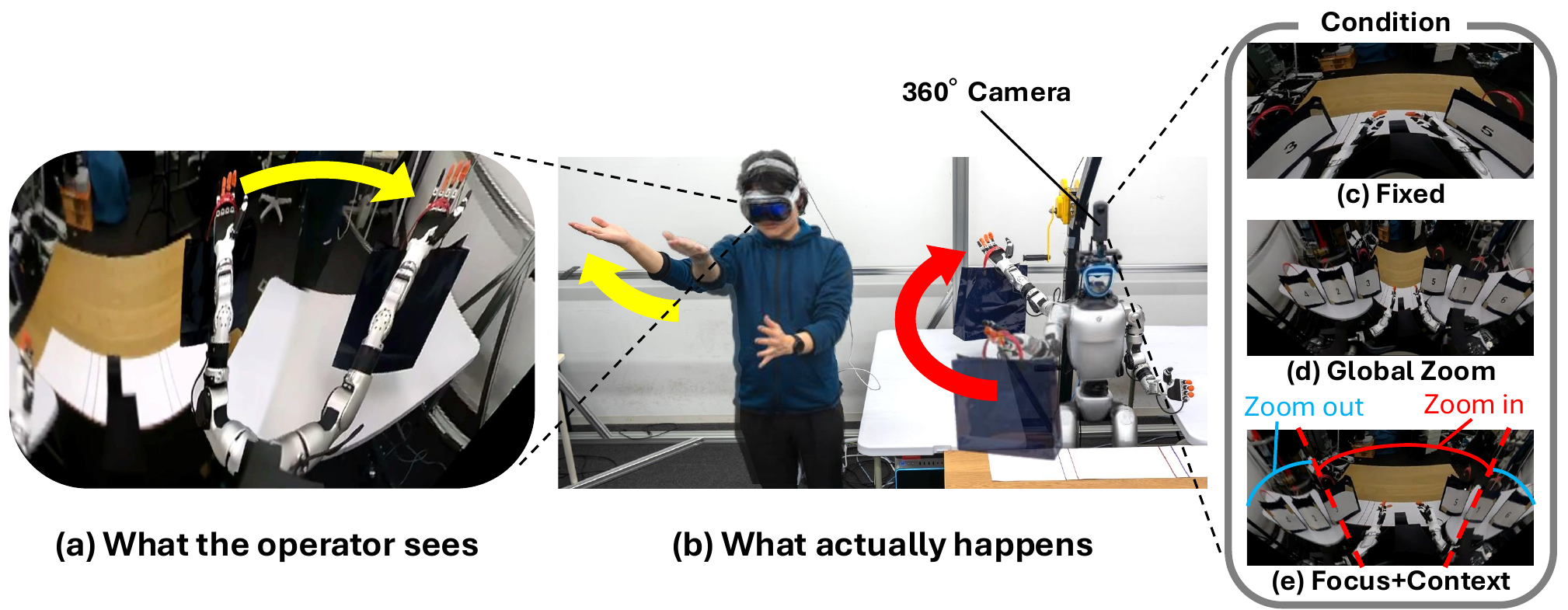}
  \caption{
    Coupled remapping brings side and rear robot targets into frontal operator space while matching displayed and commanded reach directions; panels show Fixed,  Global Zoom, and Focus+Context.
  }
    \label{fig:concept}
\end{teaserfigure}

\begin{abstract}
First-person humanoid teleoperation ties access to side and rear targets to operator rotation.
We present coupled visual--motor remapping, pairing a panoramic azimuth warp with its inverse upper-limb yaw transform.
A working prototype implements Global Zoom (GZ) and Focus+Context (F+C).
In a within-subject study (\(N=8\); 72 trials), GZ had the largest completion count (23/24; Fixed: 20/24; F+C: 21/24), whereas F+C was preferred by 6/8 participants.
Thus, completion and preference favored different mappings in this sample.
\end{abstract}

\maketitle

\section{Introduction}

Immersive humanoid teleoperation lets operators act through a remote body from its viewpoint~\cite{darvish2023survey,fritsche2015first}. First-person systems commonly preserve a one-to-one relationship among the operator's view, body motion, and robot motion. This correspondence is easy to interpret, but it also ties access to side and rear targets to large head and torso rotations. Widening the view can expose more of the workspace, but reduces local visual detail.

Panoramic warping offers another possibility: redistributing a 360$^\circ$ workspace into the operator's frontal field of view. However, visual remapping alone breaks the action correspondence that makes first-person teleoperation understandable. A lateral target may appear in front while the commanded arm direction remains lateral, leaving the operator to see and act through different spatial mappings. Rather than treating zoom as a display-only function, we treat the allocation of visual space as part of the action mapping.

Prior teleoperation systems mirror body motion, steer robot cameras, add auxiliary views, or modify control mappings~\cite{fritsche2015first,cheng2025opentelevision,iyer2025openteach,tang2025cafe}. Fisheye, peripheral-compression, and zoom interfaces redistribute visual detail~\cite{furnas1986fisheye,orlosky2014fisheye,mizukoshi2020zoom}. Most closely, Vaughan et al.\ applied focus+context visualization to mobile-robot navigation~\cite{vaughan2016look}. Our approach instead couples a single panoramic view warp with inverse upper-limb yaw remapping for humanoid manipulation, so that displayed and commanded reach directions remain aligned.

We contribute (1) this coupled visual--motor remapping principle and (2) a working humanoid teleoperation prototype implementing uniform Global Zoom (GZ) and non-uniform Focus+Context (F+C). An exploratory study compares their direction-specific task completion and overall user preference.

\section{System}

After pose conversion to a common robot-aligned frame, let \(W_c\) map robot-relative azimuth to display azimuth and let \(\theta_h,\theta_w\) denote head and wrist yaw.
Control applies

\begin{equation}
  \phi = \operatorname{wrap}(\theta_w-\theta_h),
  \qquad
  \theta_r = \theta_h + W_c^{-1}(\phi),
\end{equation}

where \(\theta_r\) is robot reach azimuth.
Rendering uses \(W_c\), and control uses \(W_c^{-1}\), geometrically matching displayed and commanded directions.

\begin{figure}[t]
  \centering
  \includegraphics[width=\columnwidth]{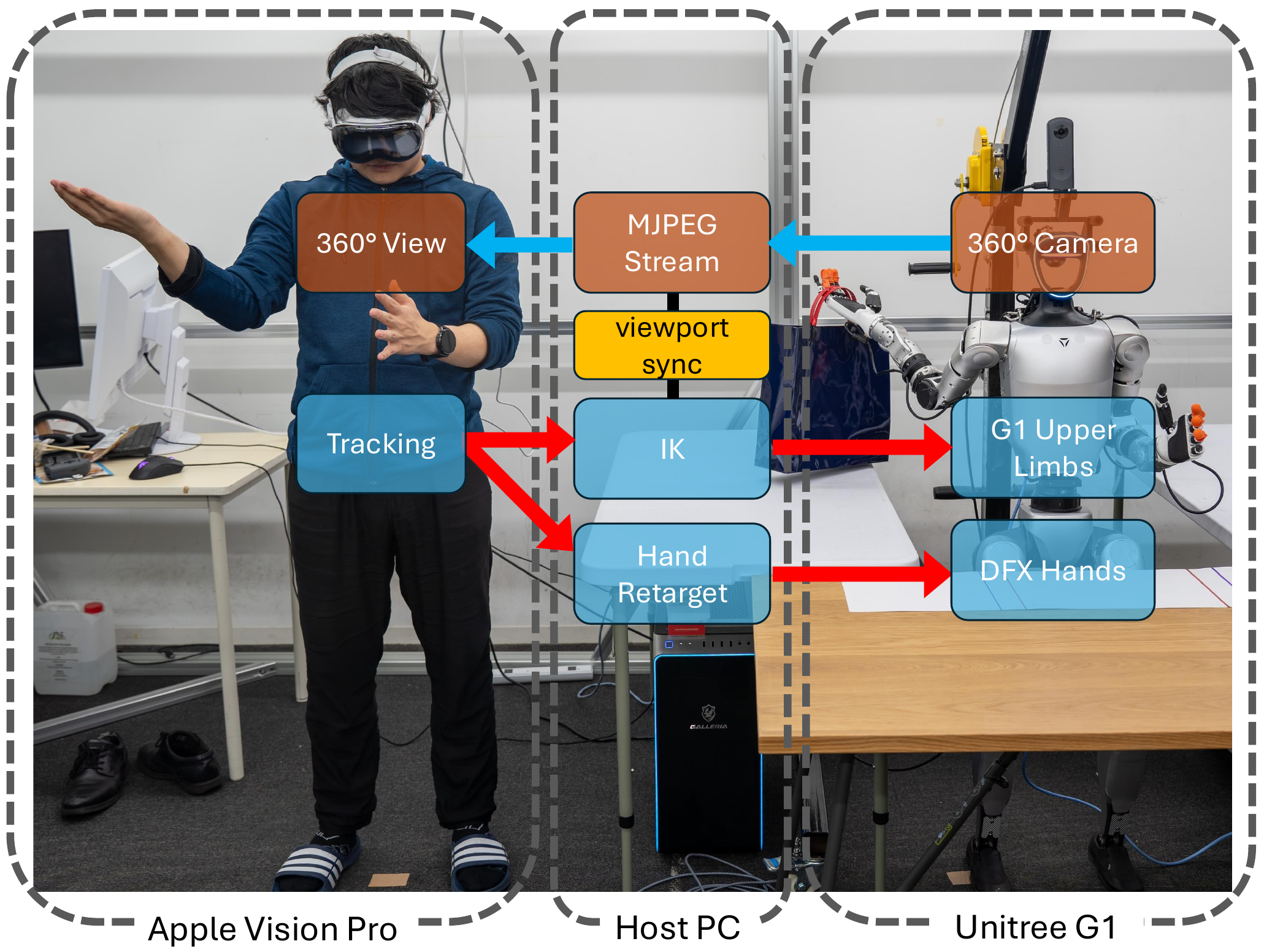}
  \caption{
    Video flows robot--host--HMD; tracking and viewport state return for inverse mapping and upper-limb control.
  }
  \label{fig:system}
\end{figure}

A RICOH THETA X on a Unitree G1 streams a monocular 360\textdegree{} view to an Apple Vision Pro.
The headset returns tracking and viewport state to a host, which remaps wrist poses before inverse kinematics (Figure~\ref{fig:system}).

\begin{figure}[t]
  \centering
  \includegraphics[width=\columnwidth]{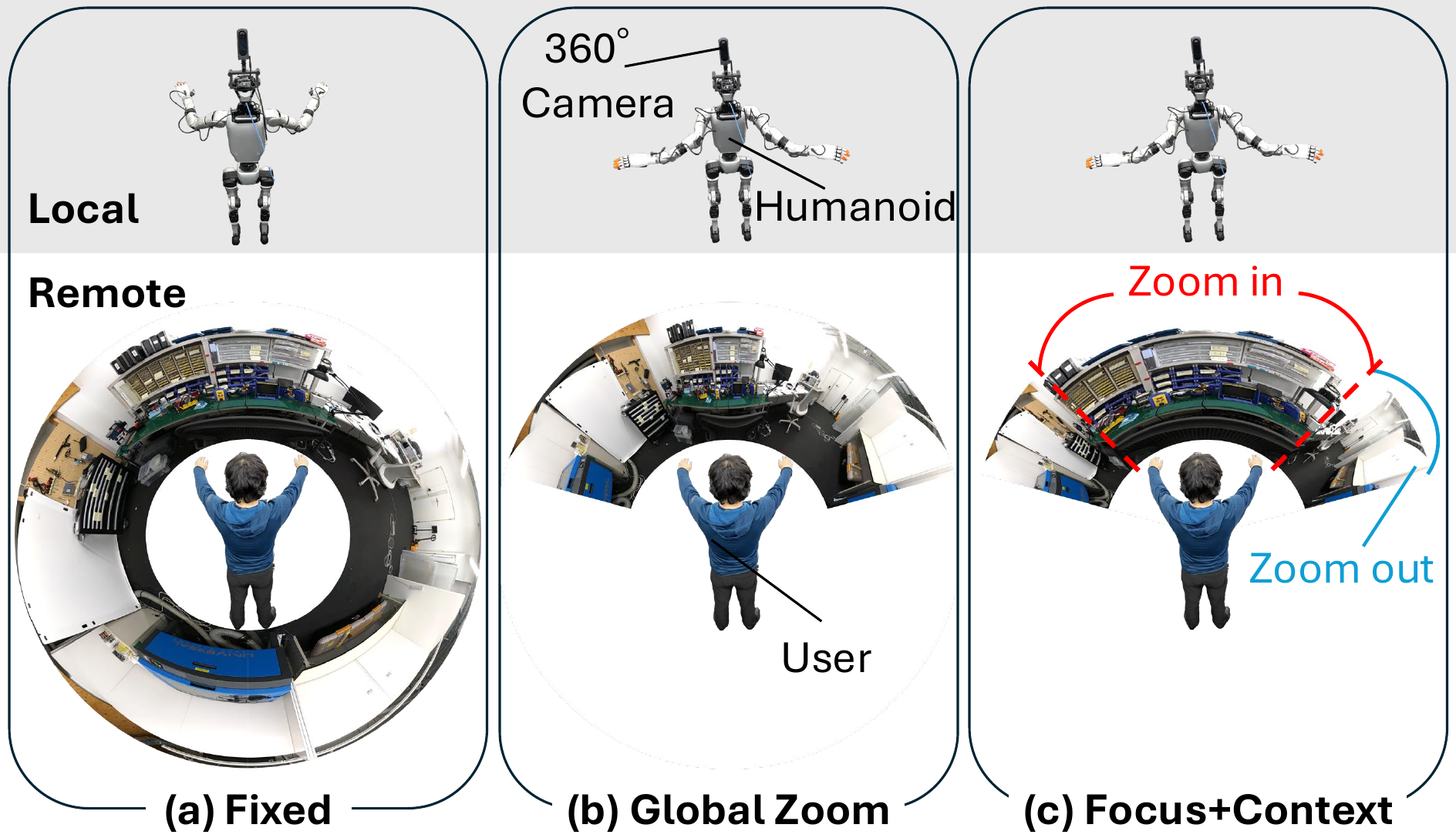}
  \caption{
    Fixed preserves identity;  Global Zoom uniformly compresses 360\textdegree{} into 180\textdegree{}; Focus+Context preserves more central detail with an inverse nonlinear reach mapping.
  }
  \label{fig:mappings}
\end{figure}

\textbf{Fixed} uses identity mappings.
\textbf{GZ} uniformly maps 360\textdegree{} into 180\textdegree{} with yaw gain 2.0.
\textbf{F+C} uses the same span but expands a central \(\pm15^\circ\) region by \(1.5\times\) and compresses the periphery, with the corresponding inverse control warp (Figure~\ref{fig:mappings}).

\section{Evaluation}

\begin{figure}[t]
  \centering
  \includegraphics[width=\columnwidth]{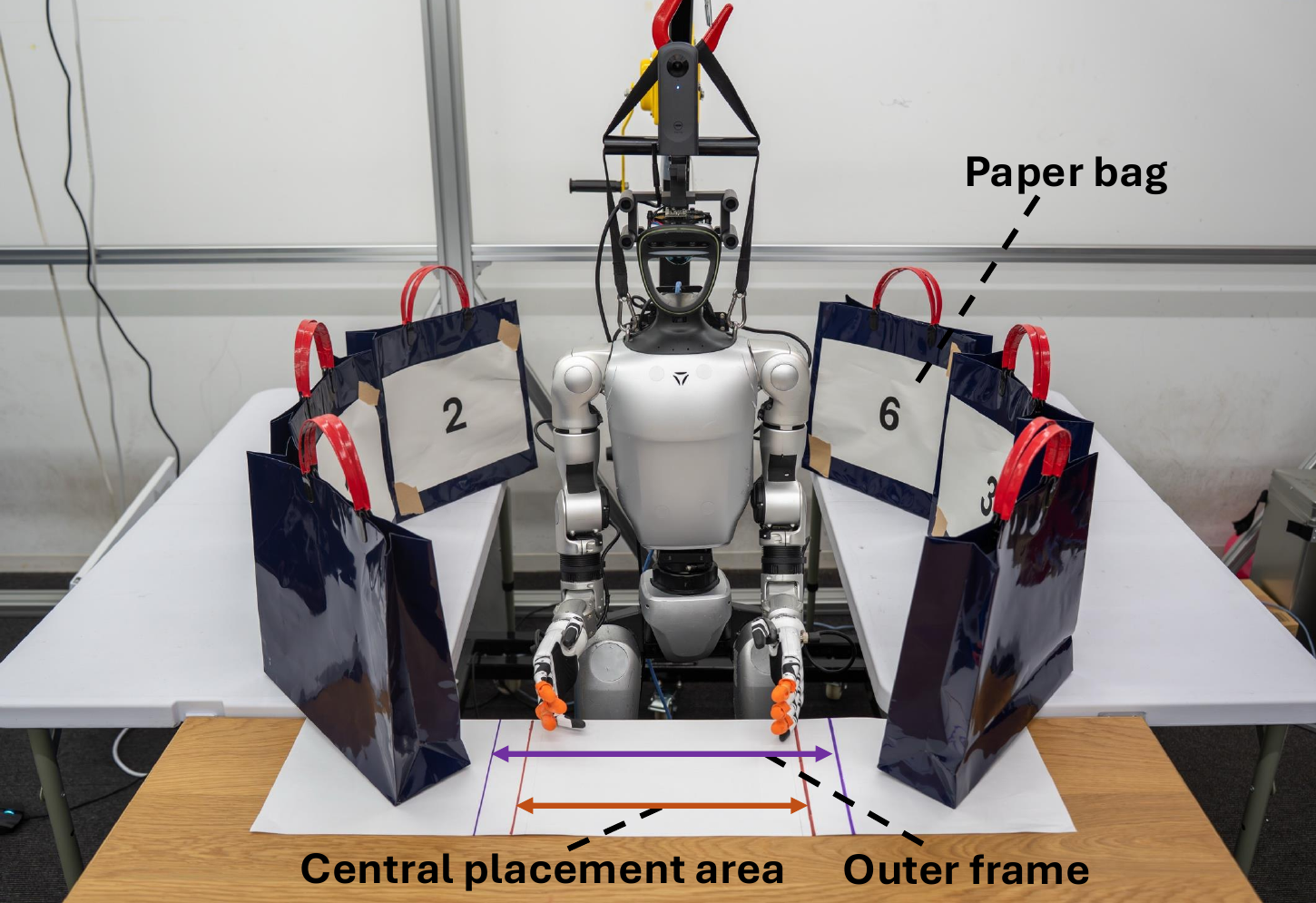}
  \caption{
    Participants moved bags from front (\(\pm30^\circ\)), side
    (\(\pm90^\circ\)), or rear (\(\pm120^\circ\)) into the central area.
    Completion required the entire bag inside the 40\,cm outer frame;
    trials stopped after five knock-overs or five minutes.
  }
  \label{fig:experiment}
\end{figure}

Eight participants performed one trial per condition and target direction
using fixed-base, upper-limb control (72 trials; condition order
counterbalanced; Figure~\ref{fig:experiment}).

\begin{table}[t]
  \centering
  \caption{Completed trials and overall-preference votes (\(N=8\)).}
  \label{tab:results}
  \footnotesize
  \setlength{\tabcolsep}{2.8pt}
  \begin{tabular}{@{}lccccc@{}}
    \toprule
    Condition & Front & Side & Rear & Total & Pref. \\
    \midrule
    Fixed & 8/8 & 8/8 & 4/8 & 20/24 & 1/8 \\
    GZ    & 8/8 & 7/8 & 8/8 & \textbf{23/24} & 1/8 \\
    F+C   & 8/8 & 8/8 & 5/8 & 21/24 & \textbf{6/8} \\
    \bottomrule
  \end{tabular}
\end{table}

GZ produced the largest completion count (23/24), whereas F+C received
six of eight preference votes. Thus, completion and preference favored
different mappings in this sample.

\section{Conclusion}

Coupled remapping makes view allocation part of the action mapping. In this sample, GZ yielded the most completed trials, whereas F+C received the most preference votes. This divergence motivates evaluating task completion and user preference as separate design criteria.

\begin{acks}
This work was supported by the JST Moonshot R\&D Program (Grant Number JPMJMS2012), the JST ASPIRE Program (Grant Number JPMJAP2503), and JST SPRING (Grant Number JPMJSP2108).
\end{acks}

\bibliographystyle{ACM-Reference-Format}
\bibliography{references}

@article{darvish2023survey,
  author  = {Darvish, Kourosh and Penco, Luigi and Ramos, Joao and Cisneros, Rafael and Pratt, Jerry and Yoshida, Eiichi and Ivaldi, Serena and Pucci, Daniele},
  title   = {Teleoperation of Humanoid Robots: A Survey},
  journal = {IEEE Transactions on Robotics},
  year    = {2023},
  volume  = {39},
  number  = {3},
  pages   = {1706--1727},
  doi     = {10.1109/TRO.2023.3236952}
}

@inproceedings{fritsche2015first,
  author    = {Fritsche, Lars and Unverzagt, Felix and Peters, Jan and Calandra, Roberto},
  title     = {First-Person Tele-Operation of a Humanoid Robot},
  booktitle = {2015 IEEE-RAS 15th International Conference on Humanoid Robots (Humanoids)},
  year      = {2015},
  pages     = {997--1002},
  doi       = {10.1109/HUMANOIDS.2015.7363475}
}

@inproceedings{cheng2025opentelevision,
  author    = {Cheng, Xuxin and Li, Jialong and Yang, Shiqi and Yang, Ge and Wang, Xiaolong},
  title     = {{Open-TeleVision}: Teleoperation with Immersive Active Visual Feedback},
  booktitle = {Proceedings of The 8th Conference on Robot Learning},
  series    = {Proceedings of Machine Learning Research},
  volume    = {270},
  pages     = {2729--2749},
  year      = {2025},
  publisher = {PMLR},
  url       = {https://proceedings.mlr.press/v270/cheng25b.html}
}

@inproceedings{iyer2025openteach,
  author    = {Iyer, Aadhithya and Peng, Zhuoran and Dai, Yinlong and Guzey, Irmak and Haldar, Siddhant and Chintala, Soumith and Pinto, Lerrel},
  title     = {{OPEN TEACH}: A Versatile Teleoperation System for Robotic Manipulation},
  booktitle = {Proceedings of The 8th Conference on Robot Learning},
  series    = {Proceedings of Machine Learning Research},
  volume    = {270},
  pages     = {2372--2395},
  year      = {2025},
  publisher = {PMLR},
  url       = {https://proceedings.mlr.press/v270/iyer25a.html}
}

@misc{tang2025cafe,
  author        = {Tang, Zixin and Chen, Yiming and Rouxel, Quentin and Li, Dianxi and Wu, Shuang and Chen, Fei},
  title         = {{CaFe-TeleVision}: A Coarse-to-Fine Teleoperation System with Immersive Situated Visualization for Enhanced Ergonomics},
  year          = {2025},
  eprint        = {2512.14270},
  archivePrefix = {arXiv},
  primaryClass  = {cs.RO},
  doi           = {10.48550/arXiv.2512.14270},
  url           = {https://arxiv.org/abs/2512.14270}
}

@inproceedings{furnas1986fisheye,
  author    = {Furnas, George W.},
  title     = {Generalized Fisheye Views},
  booktitle = {Proceedings of the SIGCHI Conference on Human Factors in Computing Systems},
  year      = {1986},
  pages     = {16--23},
  publisher = {Association for Computing Machinery},
  address   = {New York, NY, USA},
  doi       = {10.1145/22627.22342}
}

@inproceedings{orlosky2014fisheye,
  author    = {Orlosky, Jason and Wu, Qifan and Kiyokawa, Kiyoshi and Takemura, Haruo and Nitschke, Christian},
  title     = {Fisheye Vision: Peripheral Spatial Compression for Improved Field of View in Head Mounted Displays},
  booktitle = {Proceedings of the 2nd ACM Symposium on Spatial User Interaction},
  year      = {2014},
  pages     = {54--61},
  publisher = {Association for Computing Machinery},
  address   = {New York, NY, USA},
  doi       = {10.1145/2659766.2659771}
}

@inproceedings{mizukoshi2020zoom,
  author    = {Mizukoshi, Yuichi and Sato, Ryuya and Eto, Takahiro and Kamezaki, Mitsuhiro and Matsuzaka, Ayaka and Liu, Yang and Namiki, Akio and Imai, Asaki and Matsuzawa, Takashi and Hashimoto, Kenji and Takanishi, Atsuo and Iwata, Hiroyasu},
  title     = {A Low Cognitive Load and Reduced Motion Sickness Inducing Zoom Method Based on Typical Gaze Movement for Master-Slave Teleoperation Systems with HMD},
  booktitle = {2020 IEEE/SICE International Symposium on System Integration (SII)},
  year      = {2020},
  pages     = {28--33},
  doi       = {10.1109/SII46433.2020.9026260}
}

@inproceedings{vaughan2016look,
  author    = {Vaughan, Jim and Kratz, Sven and Kimber, Don},
  title     = {Look Where You're Going: Visual Interfaces for Robot Teleoperation},
  booktitle = {2016 25th IEEE International Symposium on Robot and Human Interactive Communication (RO-MAN)},
  year      = {2016},
  pages     = {273--280},
  doi       = {10.1109/ROMAN.2016.7745142}
}

\end{document}